# Direct $\ell_{2,p}$-Norm Learning for Feature Selection

**Hanyang Peng, Yong Fan**

National Laboratory of Pattern Recognition, Brainnetome centre,
Institute of Automation, Chinese Academy of Science, Beijing, China, 100190
{hanyang.peng, yfan}@nlpr.ia.ac.cn

**Abstract**

In this paper, we propose a novel sparse learning based feature selection method that directly optimizes a large margin linear classification model's sparsity with $\ell_{2,p}$-norm ($0 < p \leq 1$) subject to data-fitting constraints, rather than using the sparsity as a regularization term. To solve the direct sparsity optimization problem that is non-smooth and non-convex when $0 < p < 1$, we provide an efficient iterative algorithm with proved convergence by converting it to a convex and smooth optimization problem at every iteration step. The proposed algorithm has been evaluated based on publicly available datasets, and extensive comparison experiments have demonstrated that our algorithm could achieve feature selection performance competitive to state-of-the-art algorithms.

## 1. Introduction

Nowadays, we face to handle more and more high-dimensional data in machine learning. Since many features in real applications are not informative, they may lead to over-fitting and deteriorated learning models for small sample size problems. To improve the efficacy of machine learning models, feature selection has been an important tool(Guyon & Elisseeff, 2003).

Feature selection approaches in general can be divided into three groups: filter methods (Kira & Rendell, 1992; Lewis, 1992; Peng et al., 2005), wrapper methods (Guyon et al., 2002), and embedded methods (Cawley et al., 2006; Wang et al., 2008; Xiang et al., 2012). Filter methods use proxy measures that are independent on the learning models to rank features according to their relevancy to the learning problem. Wrapper methods directly utilize a learning model's performance to select features. Therefore, the wrapper methods typically have higher computational cost than the filter methods. Since embedded methods integrate feature selection into the model learning, they typically achieve good performance with moderate computational cost.

As an embedded method, sparse linear model based feature selection has attracted much attention, and many variants have been proposed with different sparsity regularization terms. In particular, $\ell_1$-norm regularization based algorithms, such as Lasso (Tibshirani, 1996) and sparse SVM (Bradley & Mangasarian, 1998; Wang, et al., 2008), have been demonstrated powerful in feature selection. In multi-task learning, various $\ell_{2,1}$-norm (Liu et al., 2009; Nie et al., 2010; Obozinski et al., 2006) or $\ell_{\infty,1}$-norm(Liu et al., 2009) based regularization models have been investigated for selecting features with joint sparsity cross different tasks. Moreover, group Lasso based methods (Kong & Ding, 2013; Kong et al., 2014) have also been proposed in recent years. In fact, the multi-task feature selection algorithms have close connection to group lasso based methods. Since non-convex $\ell_p$-norm or $\ell_{r,p}$-norm ($0 < p < 1$) based regularization models can yield more sparse solution with no bias than $\ell_1$-norm or $\ell_{r,1}$-norm based models (Fan & Peng 2004; Tan et al., 2014), they have gained increasing attention in recent studies (Chartrand & Staneva, 2008; Liu et al., 2007; Zhang et al., 2014).

Sparsity regularization based feature selection algorithms typically take a compromise between data-fitting loss function term and sparsity term, therefore there inevitably exists residual in the loss function that will have an impact on feature selection. In this paper, we propose a novel algorithm via directly optimizing sparsity of a linear model with a $\ell_{2,p}$-norm ($0 < p \leq 1$) based objective function, instead of adopting the sparsity as a regularization term. This direct sparsity optimization method is further enhanced by a large margin model learning technique. We provide an efficient algorithm to solve the non-convex and non-smooth optimization problem associated with the direct sparsity optimization by transforming it to a Frobenius-norm induced problem at each iteration step, which has been proved to converge to the optimal solution.

The proposed algorithm has been evaluated based on 9 publicly available datasets, and extensive comparison experiments have demonstrated that our algorithm could achieve feature selection performance competitive to state-of-the-art algorithms, including $\ell_1$-norm based Lasso (Tibshirani, 1996), $\ell_{2,1}$-norm based Robust Feature Selection (RFS) (Nie, et al., 2010; Xiang, et al., 2012), ReliefF (Kira & Rendell, 1992), and mRMR(Peng, et al., 2005).

## 2. Feature Selection via Direct Sparsity Optimization (DSO-FS)

Throughout this paper, matrices are written in bold uppercase, vectors are written in bold lowercase, and all the scalars are denoted by un-boldface letters. $I$ denotes an identity matrix and $\mathbf{1}$ denotes a vector or matrix with all the elements equal to 1. Given a matrix $A \in \mathbb{R}^{m \times n}$, the $\ell_{r,p}$-norm($r > 0, p > 0$)[1] of $A$ is defined as:

$$\|A\|_{r,p} = \left( \sum_{i=1}^{m} \left( \sum_{j=1}^{n} |A_{i,j}|^r \right)^{\frac{p}{r}} \right)^{\frac{1}{p}} = \left( \sum_{i=1}^{m} (\|a_i\|_r)^p \right)^{\frac{1}{p}}, (1)$$

where $\|a_i\|_r$ denotes $\ell_r$-norm of the $i$-th row vector of $A$.

Given $m$ training samples $\{x^i, y^i\}_{i=1}^{m}$ where $x^i \in \mathbb{R}^n$ is a data point and $y^i$ is its associated class label in $c (c \geq 2)$ classes, the multiclass classification problem can be modeled as a linear learning problem, i.e.,

$$XW = Y, \qquad (2)$$

where $X = \{x^i\}_{i=1}^{m}$, $W \in \mathbb{R}^{n \times c}$ is the weight matrix to be learned, and $Y = [f^1; \ldots; f^i; \ldots; f^n] \in \mathbb{R}^{n \times c}$ and is a class label matrix with labels rearranged using a one-versus-rest model, i.e., $f^i = [-1, \ldots, 1, \ldots, -1] \in \mathbb{R}^c$ (the $j$-th element is 1 and others are $-1$ for the $i$-th data point belonging to the $j$-th class).

To achieve feature selection, $W$ should have sparse column vectors and share joint sparsity along its row direction since each row of $W$ corresponds to the same feature. Therefore, we model the feature selection problem as[1].

$$\min_{W} \|W\|_{2,0}, \quad s.t., XW = Y. \qquad (3)$$

where $\|W\|_{2,0}$ is the number of rows in $W$ of which not all the elements are zero.

Solving the optimization problem of Eqn.(3) is NP-hard. Therefore $\ell_p$-norm ($0 < p \leq 1$) can be adopted instead, resulting in a relaxed sparsity optimization problem:

$$\min_{W} \|W\|_{2,p}, \quad s.t., XW = Y, \qquad (4)$$

where $0 < p \leq 1$.

Many studies assumed that a $\ell_{2,1}$-norm used in Eqn. (4) could lead to the sparsest solution (Liu, et al., 2009; Nie, et al., 2010; Obozinski, et al., 2006). However, such a strategy works only under certain conditions (Candes & Tao, 2005). In practice, $\ell_{2,p}$-norm($0 < p < 1$) will help get more sparse solution in most cases although it is non-convex.

It is desired that the classification model's margin between classes is as large as possible for obtaining improved generalization performance. Accordingly, the equality constraint in Eqn. (4) is relaxed to be inequality constraints, i.e.,

$$\min_{W} \|W\|_{2,p}, \quad s.t., Y \odot XW \geq \mathbf{1}, \qquad (5)$$

where $\mathbf{1} \in \mathbb{R}^{m \times c}$, and $\odot$ is a Hadamard product operator that makes a direct production within the corresponding elements of both matrices. And the optimization problem of Eqn. (5) can be formulated by introducing slack variables $E \in \mathbb{R}^{m \times c}$ whose elements have the same positive or negative sign as the corresponding elements of $Y$, as

$$\min_{W,E} \|W\|_{2,p}$$
$$s.t. \quad XW = Y + E$$
$$Y \odot E \geqslant \mathbf{0}, \qquad (6)$$

where $E \in \mathbb{R}^{m \times c}$ and $\geqslant 0$ denotes that all the elements in the matrix are greater than or equal to 0.

Our strategy to solve the optimization problem of Eqn. (6) are the following. First, we solve the linear equation $XW = Y + E$ to obtain the solution space of $W$. Then, we directly search the solution space to find a solution with the minimum of $\|W\|_{2,p}$. Note that if linear equation $XW = Y + E$ is inconsistent, especially when the number of data samples $m$ is greater than the number of features $n$, least-square solution space of the equation can be substituted. Actually, we just need to solve $XW = XX^+(Y + E)$, where $X^+$ is pseudo-inverse. This equation is compatible when $XW = Y + E$ is consistent, since on this occasion $XX^+ = I$. Gaussian Elimination is a simple and efficient way to obtain the solve space of $W$, i.e.,

$$[X \vdots XX^+(Y + E)] = [X_1 \ X_2 \vdots XX^+(Y + E)]$$
$$\xrightarrow{\text{premultipy } D} \begin{bmatrix} I & M & \vdots & N + LE \\ 0 & 0 & \vdots & 0 \end{bmatrix}, \qquad (7)$$

where the rank of $X$ is $m_0$ and $n_0 = n - m_0$, $X_1 \in \mathbb{R}^{m \times m_0}$, $X_2 \in \mathbb{R}^{m \times n_0}, I \in \mathbb{R}^{m_0 \times m_0}$, $M \in \mathbb{R}^{m_0 \times n_0}$, $N \in \mathbb{R}^{m_0 \times c}$, and $L = DXX^+ \in \mathbb{R}^{m_0 \times m_0}$. Thus, the solution space of $W$ is

$$W = PU + Q + F = \begin{bmatrix} M \\ I \end{bmatrix} U + \begin{bmatrix} N \\ 0 \end{bmatrix} + \begin{bmatrix} LE \\ 0 \end{bmatrix}, \quad (8)$$

where $P \in \mathbb{R}^{n \times n_0}$, $U \in \mathbb{R}^{n_0 \times c}$, $Q \in \mathbb{R}^{n \times c}$, $F \in \mathbb{R}^{n \times c}$, and $I \in \mathbb{R}^{n_0 \times n_0}$.

---

[1] If $0 < r < 1$ or $0 < p < 1$, $\ell_{r,p}$-norm does not satisfy triangle inequality, but it is not needed in this paper.

[1] For simplicity, the bias of the standard linear regression is absorbed into $W$ as an additional dimension with all elements equal to 1.

Finally, the optimization problem of Eqn. (5) can be reformulated as

$$\min_{U,E} \left\| PU + Q + \begin{bmatrix} LE \\ 0 \end{bmatrix} \right\|_{2,p}, \text{ s.t., } Y \odot E \geqslant 0, \quad (9)$$

where $0 < p \leq 1$.

## 3. Solution to DSO-FS

### 3.1 An iterative algorithm to DSO-FS

We propose an iterative strategy to solve the non-linear optimization problem of Eqn. (9) due to that no analytical solution is available. At each iteration step, we alternately optimize variables $U$ and $E$. An optimization problem with $\ell_{2,p}$-norm ($0 < p \leq 1$) is non-smooth and non-convex when $0 < p < 1$. To efficiently solving this problem, $\ell_{2,p}$-norm is reformulated by Frobenius-norm ($\mathcal{F}$-norm) that is smooth and convex, as

$$\|A\|_{2,p}^p = \|\Sigma A\|_F^2, \quad (10)$$

where $\Sigma$ is a diagonal matrix with the $i$-th diagonal element $\Sigma_{ii} = 1/\|a_i\|_2^{1-p/2}$, and $\|a_i\|_2$ is defined in Eqn.(1)[2]. Such a strategy was first developed in FOcal Underdetermined System Solver (FOCUSS) (Gorodnitsky & Rao, 1997), and has been adopted in various $\ell_{2,1}$-norm regularization based feature selection algorithms (Hou et al., 2011; Nie, et al., 2010; Xiang, et al., 2012; Yi Yang et al., 2011).

Our solution to DSO-FS is summarized in Algorithm 1.

**Algorithm 1** Feature Selection via Direct Sparsity Optimization (DSO-FS)

**Input**: data points $\{x^i\}_{i=1}^m$ ($x^i \in \mathbb{R}^n$) and their corresponding label $\{y^i\}_{i=1}^m$; norm power $p$; number of features $d$ to be selected.

Construct $X$ and $Y$ according to Eqn. (2), and $L$, $M$, $N$, $P$, and $Q$ according to Eqn.(7) and Eqn.(8).

Set $k = 0$ and initialize $\Sigma^0 \in \mathbb{R}^{n \times n}$ with an identity matrix.

**repeat**

$G = Q + \begin{bmatrix} LE^k \\ 0 \end{bmatrix}$

$U^{k+1} = \arg\min_U \|\Sigma^k(PU + G)\|_F^2$

$V^k = M\widetilde{U}^{k+1} + N + LE^k$, where $\widetilde{U}^{k+1}$ is the first $m_0$ row vectors of $U^{k+1}$

Update diagonal matrix $\Lambda^k$, where the $i$-th diagonal element is $\frac{1}{\|v_i^k\|_2^{1-p/2}}$

$H = M\widetilde{U}^k + N$

$E^{k+1} = \arg\min_E \|\Lambda^k(LE + H)\|_F^2$, s.t. $Y \odot E \geqslant 0$

Update $k = k + 1$, $W^k = PU^k + Q + \begin{bmatrix} LE^k \\ 0 \end{bmatrix}$

Update $\Sigma^k$, where the $i$-th diagonal element is $\frac{1}{\|w_i^k\|_2^{1-p/2}}$

**until** *convergence*

Sort all features according to $\|w_i\|_2$ and select the top largest $d$ features.

Note that at each iteration step, for solving the problem $U^{k+1} = \arg\min_U \|\Sigma^k(PU + G)\|_F^2$, an analytical solution is available, i.e.,

$$U^{k+1} = (P^t S P)^{-1} P^t S G. \quad (11)$$

where $S = (\Sigma^k)^2$

According to Eqn. (8), Eqn. (11) can be reformulated as

$$U^{k+1} = (M^t S_1 M + S_2)^{-1} M^t S_1 K, \quad (12)$$

where $S_1 \in \mathbb{R}^{m_0 \times m_0}$ is a diagonal matrix, of which the diagonal elements are the first $m_0$ elements of $S$, $S_2 \in \mathbb{R}^{n_0 \times n_0}$ is a diagonal matrix, of which the diagonal elements are the last $n_0$ elements of $S$, and $K = N + LE^k (\in \mathbb{R}^{m_0 \times c})$.

If $m_0 < n_0$, we can reduce the optimization problem's complexity with a simple matrix operation

$$U^{k+1} = T(MT + I)^{-1} K, \quad (13)$$

where $T = S_2^{-1} M^t S_1$ and $I \in \mathbb{R}^{m_0 \times m_0}$.

For feature selection problems with the number of samples less than the number of features, i.e., $m_0 < m \leq n_0$, updating Eqn. (13) has reduced computational cost than Eqn. (11). It is worth noting that it is not necessary to directly calculate $C = (MT + I)^{-1} K$, which is computationally more expensive than solving the following linear equation

---

[2] Actually, $\Sigma = \Pi^{-1}$, where $\Pi$ is a diagonal matrix with the $i$-th diagonal element $\Pi_{ii} = \|a_i\|_2^{1-p/2}$. If any row vector $a_i = 0$, we define $\Sigma = \Pi^+$.

$$(MT + I)C = K. \quad (14)$$

Although no analytical solution is not available for $E^{k+1} = \arg\min_E \|\Lambda^k(LE + H)\|_F^2$ ( s.t. $Y \odot E \geq 0$ ), the problem itself is a smooth and convex optimization problem that can be efficiently solved by existing tools, such as CVX(CVX Research, 2011).

## 3.2 Convergence proof

Algorithm 1 makes $\|W\|_{2,p}$ monotonically decreasing with every iteration step and the optimization finally converges to the optimum.

**Lemma 1.** *Given any two vectors $a$ and $b$, we have*

$$(1-\theta)\|a\|_2^2 + \theta\|b\|_2^2 \geq \|a\|_2^{2-2\theta}\|b\|_2^{2\theta}, \quad (15)$$

*where $0 < \theta < 1$ and the equality holds if and only if $a = b$.*

*Proof.* Since $\ln(x^2)$ is concave, we have

$$\ln\big((1-\theta)x_1^2 + \theta x_2^2\big) \geq (1-\theta)\ln(x_1^2) + \theta\ln(x_2^2), (16)$$

where $0 < \theta < 1$. The equality holds if and only if $x_1 = x_2$, indicating that

$$(1-\theta)x_1^2 + \theta x_2^2 \geq x_1^{2-2\theta}x_2^{2\theta}. \quad (17).$$

Then we have

$$(1-\theta)\|a\|_2^2 + \theta\|b\|_2^2 \geq \|a\|_2^{2-2\theta}\|b\|_2^{2\theta}, \quad (18)$$

where the equality holds if and only if $a = b$. □

**Lemma 2.** *Given an optimization problem:*

$$\min_Z \| S\,\Phi(Z) \|_F^2, \quad s.t.\ Z \in \mathcal{F}, \quad (19)$$

*where $\Phi(Z)$ is a function of $Z$, $\mathcal{F}$ is the feasible region, and $S$ is a diagonal matrix whose $i$-th diagonal element is $1/\|\Phi(Z_0)_i\|_2^{1-p/2}$ ($Z_0$ could be any object in $\mathcal{F}$, $\Phi(Z_0)_i$ is the $i$-th row vector of $\Phi(Z_0)$ and $0 < p \leq 2$ ), we have that*

$$\|\Phi(Z^*)\|_{2,p} \leq \|\Phi(Z_0)\|_{2,p}, \quad (20)$$

*where $Z^*$ is optimal solution of the above optimization problem Eqn. (19) and the equality holds if and only if $\Phi(Z^*) = \Phi(Z_0)$*

*Proof.* Since $Z^*$ is the optimal solution, we have

$$\| S\,\Phi(Z^*) \|_F^2 \leq \| S\,\Phi(Z_0) \|_F^2. \quad (21)$$

Then

$$\sum_i \frac{\|\Phi(Z^*)_i\|_2^2}{\|\Phi(Z_0)_i\|_2^{2-p}} \leq \sum_i \|\Phi(Z_0)_i\|_2^p. \quad (22)$$

where $\Phi(Z_0)_i$ and $\Phi(Z^*)_i$ are the $i$-th row vector of $\Phi(Z_0)$ and $\Phi(Z^*)$, respectively.

According to Lemma 1, we have

$$\left(1 - \frac{p}{2}\right)\|\Phi(Z_0)_i\|_2^2 + \frac{p}{2}\|\Phi(Z^*)_i\|_2^2$$
$$\geq \|\Phi(Z_0)_i\|_2^{2-p}\|\Phi(Z^*)_i\|_2^p. \quad (23)$$

Then dividing the both sides by $\|\Phi(Z_0)_i\|_2^{2-p}$, we have

$$\|\Phi(Z^*)_i\|_2^p \leq \left(1 - \frac{p}{2}\right)\|\Phi(Z_0)_i\|_2^p$$
$$+ \frac{p}{2}\frac{\|\Phi(Z^*)_i\|_2^2}{\|\Phi(Z_0)_i\|_2^{2-p}}. \quad (24)$$

It indicates that

$$\sum_i \|\Phi(Z^*)_i\|_2^s \leq \left(1 - \frac{p}{2}\right)\sum_i \|\Phi(Z_0)_i\|_2^p$$
$$+ \frac{p}{2}\sum_i \frac{\|\Phi(Z^*)_i\|_2^2}{\|\Phi(Z_0)_i\|_2^{2-p}}. \quad (25)$$

Combining (22) and (25), we obtain

$$\sum_i \|\Phi(Z^*)_i\|_2^p \leq \sum_i \|\Phi(Z_0)_i\|_2^p. \quad (26)$$

Therefore,

$$\|\Phi(Z^*)\|_{2,p} \leq \|\Phi(Z_0)\|_{2,p}, \quad (27)$$

where the equality holds if and only if $\Phi(Z^*) = \Phi(Z_0)$.□

**Theorem 1.** *Algorithm 1 makes the objective function of Eqn.(9) decreasing at each iteration step and the solution converges to the optimum.*

*Proof.* Supposing we have obtained the solution $U^k$, $E^k$, and the objective function $W^k$ at the $(k+1)$-th iteration step, we solve the optimization problem $\min_U \|\Sigma^k(PU + G)\|_F^2$ to obtain $U^{k+1}$ by fixing $E^k$. According to Lemma 2, we have

$$\|V^k\|_{2,s} = \left\| PU^{k+1} + Q + \begin{bmatrix} LE^k \\ 0 \end{bmatrix} \right\|_{2,p}$$
$$\leq \|W^k\|_{2,p}. \quad (28)$$

Then we fix $U^{k+1}$, and solve the optimization problem $\min_E \|\Lambda^k(LE + H)\|_F^2$ to obtain $E^{k+1}$.

According to Lemma 2, we have

$$\|W^{k+1}\|_{2,p} = \left\|PU^{k+1} + Q + \begin{bmatrix} LE^{k+1} \\ 0 \end{bmatrix}\right\|$$
$$\leq \|V^k\|_{2,p}. \quad (29)$$

Combining (28) and (29), we obtain

$$\|W^{k+1}\|_{2,p} \leq \|W^k\|_{2,p}. \quad (30)$$

Thus at each iteration step, the objective function $\|W\|_{2,p}$ decreases. Because of the lower bound of $\|W\|_{2,p}$ is limited, the optimization will converge with iterations. □

When $p = 1$, Eqn. (9) is a convex optimization problem, hence the solution of Eqn. (9) obtained by using Algorithm 1 may be the global optimum. When $0 < p < 1$, it may converge to a local optimum. However it usually yields more sparse solutions than that obtained with $p = 1$. Moreover, according to Lemma 2, Algorithm 1 can also solve $\ell_{2,p}$-norm($1 < p < 2$) based problems.

It is worth noting that Lemma 2 works for sparsity regularization based feature selection algorithms, *i.e.*,

$$\min_{Z} \sum_i \|A_i Z + B_i\|_{2,p} + \lambda \|Z\|_{2,p}, \quad (31)$$

where $\lambda$ is the regularization coefficient.

Since Eqn.(31) can be reformulated as

$$\min_{Z} \|[(A_1 Z + B_1)^t, \ldots, (A_i Z + B_i)^t, \ldots, \lambda Z^t]\|_{2,p}. \quad (32)$$

The Robust Feature Selection (RFS) (Nie, et al., 2010; Xiang, et al., 2012) is a special case of Eqn.(32) when $i = 1$ and $p = 1$.

## 4. Experiments

### 4.1 Experimental datasets and settings

The proposed algorithm has been evaluated based on 9 publicly available datasets, as summarized in Table 1. In particular, 3 datasets were obtained from UCI, including ISOLET, SEMEION and GISETTE [3]. ISOLET is a speech recognition data set with 7797 samples in 26 classes, and each sample has 617 features. SEMEION contains 1593 handwritten images from ~80 persons, stretched in a rectangular box of $16 \times 16$ with a gray scale of 256 values; GISETTE contains 2 confusable handwritten digits: 4 and 9 with 7000 samples and 5000 features. For these three datasets, the number of features are less than the number of their samples. Three datasets were obtained from Bio-microarray datasets, including LUNG, CLL-SUB-111, and TOX-171. For LUNG, genes with standard deviations smaller than 50 expression units are removed and 3312 genes are reserved (Cai et al., 2007). CLL-SUB-111 and TOX-171 are obtained from feature selection @ ASU[5], which has total 111 samples and 171 samples with 11340 features and 5748 feature, respectively. Our algorithm has also been validated based on 3 image datasets, including UMIST, AR and ORL. In particular, UMIST[6] includes face images from 20 different persons, and the size of each image is $56 \times 46$. AR[7] has 130 samples with 2400 features. ORL[8] includes 400 samples with $92 \times 112$ pixels as features.

*Table 1.* Datasets used in experiments

| Data Sets | #Classes | #Features | #Samples |
|---|---|---|---|
| ISOLET | 26 | 617 | 7797 |
| SEMEION | 10 | 256 | 1593 |
| GISETTE | 2 | 5000 | 7000 |
| LUNG | 5 | 3312 | 203 |
| CLL-SUB-111 | 3 | 11340 | 111 |
| TOX-171 | 4 | 5748 | 171 |
| UMIST | 20 | 2576 | 575 |
| AR | 10 | 2400 | 130 |
| ORL | 40 | 10304 | 400 |

We compared our methods with two sparsity regularization based feature selection methods, including $\ell_1$-norm based Lasso (Tibshirani, 1996) and $\ell_{2,1}$-norm based Robust Feature Selection (RFS) (Nie, et al., 2010; Xiang, et al., 2012), where the loss function is also in $\ell_{2,1}$-norm form for rejecting outliers. Moreover, we also compared our algorithm with well-known filter feature selection methods, including ReliefF (Kira & Rendell, 1992) and mRMR (Peng, et al., 2005).

Classification accuracy was used to evaluate the feature selection method. Particularly, linear SVM (Chang & Lin, 2011) was chosen to build classifiers based on the selected features. The parameter $C$ of linear SVM classifiers were tuned using a cross-validation strategy by searching a candidate set $[10^{-4}, 10^{-3}, 10^{-2}, 10^{-1}, 1, 10^1, 10^2]$. The regularized parameter $\lambda$ in Lasso and RFS were tuned using the same cross-validation strategy by searching a candidate set $[10^{-3}, 10^{-2}, 10^{-1}, 1, 10^1, 10^2, 10^3]$.

In our experiments, we first normalized all the data to have 0 mean and unit standard deviation for each feature. 10 trials were performed on each dataset. In each trial, the samples of each dataset were randomly spitted into training and testing subsets with a ration of 6:4. For tuning parameters, a 3-fold was used for datasets with less than 200 training samples, and a 8-fold cross-validation was used for other datasets.

---

[3] Available at https://archive.ics.uci.edu/ml/index.html
[5] Available at http://featureselection.asu.edu/datasets.php
[6] Available at http://images.ee.umist.ac.uk/danny/database html
[7] Available at http://featureselection.asu.edu/datasets.php
[8] Available at http://www.cl.cam.ac.uk/Research/DTG/ at archive: pub/data/att_faces.zip

## 4.2 Effect of parameter $p$ on the classification performance of the selected

The parameter $p$ of DSO-FS may have a direct impact

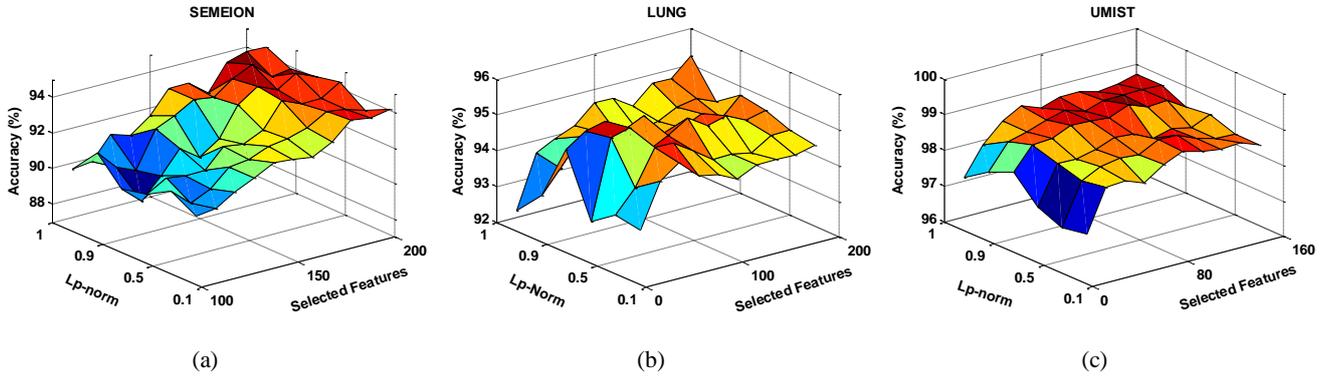

*Figure 1.* Classification accuracy with different numbers of features selected with different values of $p$. The results shown were obtained based on (a) SEMEION, (b) LUNG, and (c)UMIST.

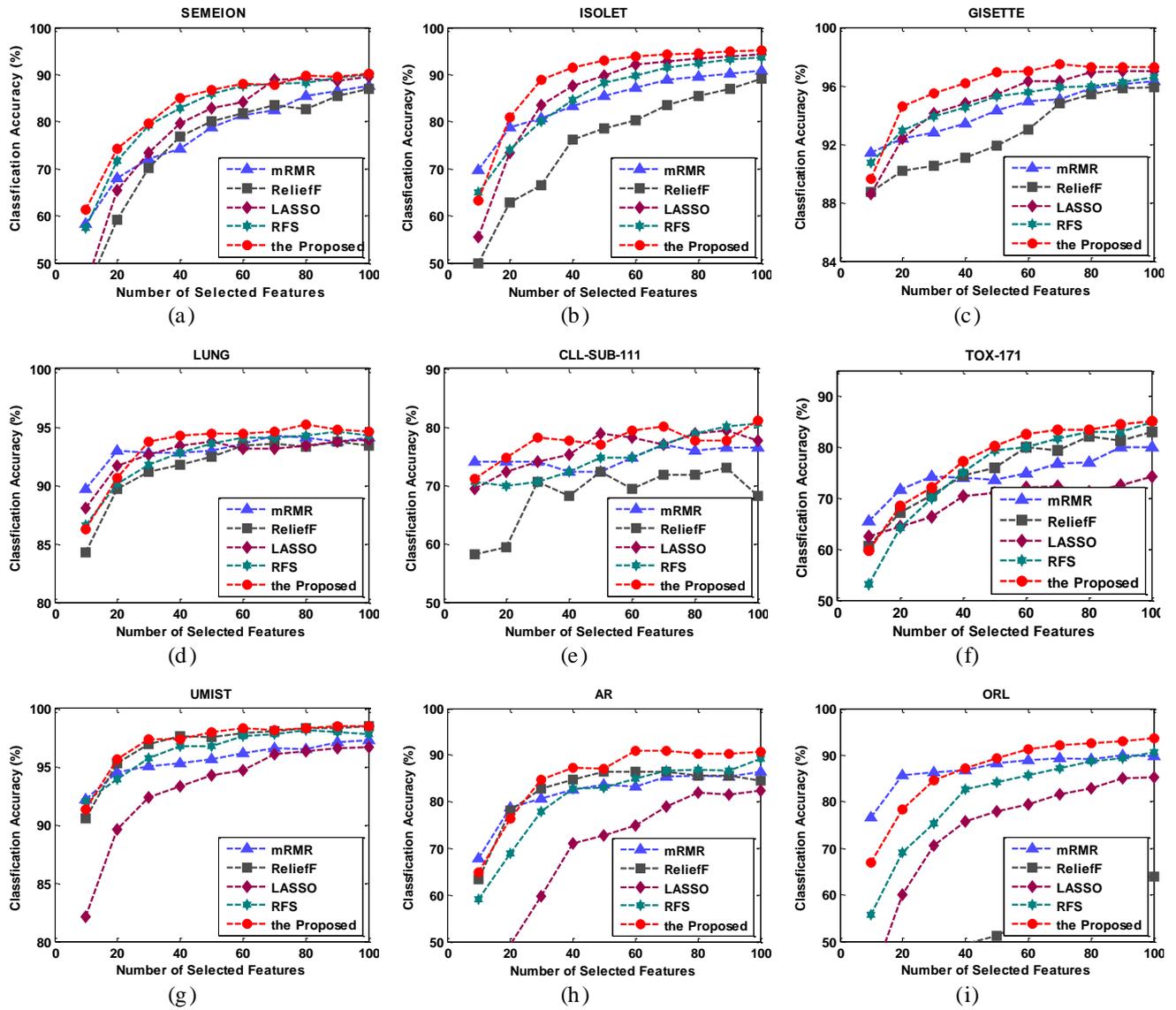

*Figure 2.* Average classification accuracy of 10 trials for classifiers built on the selected top 100 features by different algorithms. The results shown were obtained based on (a)SEMEION, (b)ISOLET, (c)GISETTE, (d)LUNG, (e)TOX-171,

(f)CLL-SUB-111, (g)UMIST, (h)AR, and (i)ORL.

Table 2. Mean and standard deviation of the classification Accuracy (%) of Linear-SVM built on the selected top 100 features by different algorithms for datasets: SEMEION, ISOLET, GISETTE, LUNG, TOX-171, CLL-SUB-111, UMIST, AR, and ORL

|  | mRMR | ReliefF | Lasso | RFS | DLO-FS |
|---|---|---|---|---|---|
| SEMEION | 87.58±1.34 | 87.04±0.77 | 89.50±0.78 | 89.92±1.62 | **90.18±0.39** |
| ISOLET | 90.82±0.72 | 89.13±0.53 | 94.25±0.69 | 93.72±0.55 | **95.23±0.44** |
| GISETTE | 96.30±0.58 | 95.90±0.87 | 95.88±1.22 | 97.0±1.12 | **97.30±1.01** |
| LUNG | 94.10±2.22 | 93.44±3.28 | 93.93±2.32 | 94.26±1.83 | **94.59±1.95** |
| CLL-SUB-111 | 76.47±3.22 | 68.24±7.30 | 77.65±3.98 | 80.59±3.99 | **81.18±3.99** |
| TOX-171 | 80.00±5.01 | 82.90±5.18 | 74.20±4.39 | 84.78±3.74 | **85.07±4.05** |
| UMIST | 97.26±0.91 | **98.48±1.12** | 96.70±1.53 | 97.83±1.54 | **98.48±1.29** |
| AR | 86.35±6.11 | 84.42±5.12 | 82.31±6.25 | 89.23±3.67 | **90.77±3.53** |
| ORL | 89.67±4.82 | 63.83±5.58 | 85.33±4.32 | 90.67±3.55 | **93.58±2.55** |

features. To investigate how the classification performance is affected by $p$, we performed experiments based on datasets UMIST, LUNG and SEMEION. We obtained different solutions of $W$ with different setting of $p \in [0.1, 0.3, 0.5, 0.7, 0.9, 1]$ using Algorithm 1 over the above 3 datasets, and then selected top ranked features according to $\ell_2$-norm $\|w_i\|_2$ to build classifiers. Figure 1 shows the classification accuracy with varying numbers of the selected features and different values of parameter $p$. The results shown in Figure 1 indicated that $p$ played an important role in the classification performance.

### 4.3 Comparisons with state-of-the-art methods

We compared the proposed method with mRMR, ReliefF, Lasso, RFS based on 9 datasets detailed in Table 1. Since the classification performance of linear SVM classifiers built on the features selected was hinged on the parameter $p$, we used a cross-validation strategy to select an optimal value from [0.1, 0.3, 0.5, 0.7, 0.9, 1].

Figure 2 shows the average classification performance of classifiers built on features selected by different methods in 10 trials. In particular, the average classification accuracy is shown as a function of the number of features used in the classification model in Figure 2. Compared with other methods, the proposed method achieved higher mean classification accuracy on most datasets, indicating that our method had overall better performance than other algorithms.

Table 2 summarizes mean and standard deviation of the classification rates in 10 trails for classifiers built on the top 100 features selected by mRMR, ReliefF, Lasso, RFS and our algorithm. These results demonstrate our algorithm had the best classification accuracy on all the 9 datasets.

### 5. Conclusions and Discussions

In this paper, a new feature selection algorithm via direct sparsity optimization was proposed. Different from the sparse regularization based algorithms, our method directly optimizes a large margin linear classification model's sparsity with $\ell_{2,p}$-norm ($0 < p \leq 1$) subject to a data-fitting constraint. We also proposed an efficient algorithm to solve the non-convex ($0 < p < 1$) and non-smooth optimization problem associated with the optimization problem. The convergence of the proposed algorithm has been rigorously proved. Extensive experiments based on 9 datasets has demonstrated that the proposed method could achieve better than 4 state-of-the-art feature selection algorithms. Our algorithm can be easily extended for solving other sparsity regularization algorithms. In particular, our algorithm could be used to obtain proximal solutions of $\ell_0$ and $\ell_{2,0}$ based optimization problems subject to linear constraints by setting $p$ close to 0.